%% file: samplepaper.tex
\documentclass[runningheads]{llncs}
\usepackage[T1]{fontenc}
\usepackage{latexsym}
\usepackage{amssymb}
\usepackage{amsmath}
\usepackage{booktabs}
\usepackage{enumitem}
\usepackage{graphicx}
\usepackage{color}
\usepackage{tabularx, array} % In preamble
\usepackage{float} 
\begin{document}
\title{Towards Inclusive NLP: Assessing Compressed Multilingual Transformers across Diverse Language Benchmarks}

\titlerunning{Towards Inclusive NLP}
\author{Maitha Alshehhi, Ahmed Sharshar, Mohsen Guizani}
\authorrunning{Maitha Alshehhi et al.}
\institute{Mohamed bin Zayed University of Artificial Intelligence, Abu Dhabi, UAE 
\email{firstname.lastname@mbzuai.ac.ae}}

\maketitle              % typeset the header of the contribution

\begin{abstract} 
Although LLMs have attained significant success in high-resource languages, their capacity in low-resource linguistic environments like Kannada and Arabic is not yet fully understood. 
This work benchmarking the performance of multilingual and monolingual Large Language Models (LLMs) across Arabic, English, and Indic languages, with particular emphasis on the effects of model compression strategies such as pruning and quantization. Findings shows significant performance differences driven by linguistic diversity and resource availability on SOTA LLMS as BLOOMZ, AceGPT, Jais, LLaMA-2, XGLM, and AraGPT2. We find that multilingual versions of the model outperform their language-specific counterparts across the board, indicating substantial cross-lingual transfer benefits. Quantization (4-bit and 8-bit) is effective in maintaining model accuracy while promoting efficiency, but aggressive pruning significantly compromises performance, especially in bigger models. Our findings pinpoint key strategies to construct scalable and fair multilingual NLP solutions and underscore the need for interventions to address hallucination and generalization errors in the low-resource setting. \footnote{ Corresponding author: Maitha Alshehhi {Maitha.alshehhi@mbzuai.ac.ae}}

\end{abstract}

\section{Introduction}

Large Language Models (LLMs) have advanced to enable transformative applications across diverse sectors. In multilingual and culturally rich environments, they enhance communication, support inclusive language technologies, and enable multilingual content generation \cite{nafea2024araoffence}. LLMs are driving innovation in fields such as healthcare, education, finance, robotics, and customer service. In finance, they aid sentiment analysis, forecasting, and portfolio optimization by processing vast unstructured data \cite{Sindhu2024evolutionllm}. In education, they support personalized learning, tutoring, and automated content creation \cite{chu2025llm}. In healthcare, they assist with diagnosis, treatment planning, and clinical decision-making \cite{nazi2024large}. Businesses benefit from LLMs through efficient document processing, summarization, chatbot development, and multilingual support \cite{cheung2024reality}.

While these advancements have demonstrated great promise, their benefits remain disproportionately centered on English and other high-resource languages. Many low-resource languages, including Arabic and Kannada, are still underrepresented in training data \cite{Hagos2024}. This underrepresentation limits the generalization capacity of LLMs and leads to higher error rates and hallucination tendencies when applied in these contexts. These issues contribute to a growing digital divide, preventing equitable access to high-quality language technologies.

Arabic and Indic languages, particularly Kannada, pose significant modeling difficulties due to their complex morphology, complicated syntax, and broad geographical variation. Poorer performance on reasoning and comprehension tasks also arise from the absence of great corpora and benchmark datasets \cite{hasan2024large,dey2024better}. These limitations hinder the successful application of LLMs to real-world tasks such as legal interpretation, content moderation on social media, and analysis of local news. It is highly probable that there will be biased or erroneous results in low-resource languages without improved cross-lingual robustness. Evaluation of models in such languages entails complex comprehension of linguistic idiosyncrasies combined with data scarcity. Efficiency is also a relevant consideration because most real-world applications need to operate under stringent hardware and latency limitations. Creating robust and equitable NLP models that generalize across different linguistic contexts depends on addressing these issues.

In this paper, we rigorously compare multilingual and monolingual LLMs on Arabic, English, and Kannada languages under standard benchmarks. Moreover, we measure the impact of compression methods—namely, pruning and quantization—on model quality, emphasizing accuracy, generalization power, and trustability under high-resource and low-resource settings. Through our large-scale experimentation, we shed light on the crucial factors that influence model size and compression methodology on the reliability and efficacy of multilingual linguistic models, thus informing the development of scalable, fair, and resilient NLP systems. Our major contributions are listed below:

\begin{itemize}

\item \textbf{Comprehensive Multilingual Benchmarking:} We provide rigorous cross-lingual evaluations of LLMs employing benchmarks such as ArabicMMLU, EnglishMMLU, and Kannada-ARC-C-2.5K.
\item \textbf{Detailed Analysis of Compression Strategies:} With a systematic evaluation of the effects of quantization and pruning on accuracy and confidence, we uncover thresholds that are pivotal to maintaining performance.
\item \textbf{Language-Specific Performance Insights:} The results of our work identify crucial Arabic-specific model limitations (e.g., AraGPT2), pointing to the relative strengths of multilingual models.
\item \textbf{Future Directions:} Our findings pave the way for a deeper understanding of hallucination patterns in LLMs, particularly in low-resource language settings. This opens new research avenues for developing more robust and trustworthy language models through targeted mitigation strategies, ultimately enabling safer deployment in real-world, multilingual applications.

% \item \textbf{Diagnostic Indicators of Trustworthiness:} A clear link is created between declining precision and confidence loss under compression, offering actionable diagnostics to validate the reliability of models after compression. Leadership
\end{itemize}

\section{Related Work}

\subsection{Addressing Low-Resource Languages}

Tackling the challenges of developing Arabic LLMs requires a multifaceted strategy, with a major emphasis on alleviating Arabic data scarcity through approaches like synthetic data generation, back-translation, and curated corpora. Models such as Jais, Falcon, and AraGPT have each made important strides: Jais, a 13-billion-parameter bilingual model, employs cross-lingual transfer learning but still falls short of GPT-3.5-turbo in trustworthiness evaluations \cite{sengupta2023jais}; Falcon, despite its massive 180B-parameter version trained on over 3.5 trillion tokens, achieves less than 50\% accuracy on ArabicMMLU, illustrating that scale alone is not sufficient for Arabic proficiency \cite{koto2024arabicmmlu}; and AraGPT, which leverages synthetic data techniques, has improved performance in Arabic sentiment classification but continues to struggle with dialectal comprehension \cite{refai2023data}. Together, these models showcase a range of strategies such as cross-lingual fine-tuning, synthetic data augmentation, and dialectal adaptation, but persistent challenges in trustworthiness, reasoning, and dialectal coverage indicate that significant gains will require further investment in data quality, domain-specific training, and dialect-focused resources [45].

\subsection{Benchmarks for LLM Evaluation}

MMLU-based benchmarks have become fundamental for evaluating LLMs across languages and subject areas, measuring reasoning, factual accuracy, and subject-specific skills. ArabicMMLU is widely used to assess Arabic-language tasks, with studies showing that multilingual models can perform competitively with monolingual ones but still show inconsistencies in reasoning and literacy tasks \cite{boughorbel2023analyzing}. However, ArabicMMLU has limitations, notably its emphasis on Modern Standard Arabic and its limited alignment with real-world applications, prompting the development of more comprehensive resources such as ArabLegalEval \cite{hijazi2024arablegaleval} and the AlGhafa Evaluation Benchmark \cite{almazrouei2023alghafa}, which better capture dialectal diversity and specialized domains.

Similarly, EnglishMMLU is a standard for assessing LLM reasoning and factual knowledge across a wide range of topics \cite{plaza2024spanish}, but faces challenges like shortcut learning, mislabeled answers, and translation artifacts that limit its real-world relevance \cite{taghanaki2024mmlu}. While models like GPT-4 and Claude consistently achieve top performance, the benchmark’s reliance on translated content often underestimates non-English model capabilities \cite{plaza2024spanish}. In low-resource settings, Indic benchmarks such as Kannada-ARC-C-2.5K are critical for evaluating LLMs \cite{singh2024indicgenbench}, although models continue to lag behind English performance due to issues like translation dependence, data imbalance, and a shortage of high-quality native datasets \cite{narayanan2024suvach}, highlighting the pressing need for more robust multilingual resources.

\subsection{Hallucination Detection \& Mitigation Strategies}

Hallucination in Arabic LLMs occurs more frequently than in English models, largely due to the scarcity of high-quality Arabic data, the complexity of dialectal variation, and the absence of structured knowledge sources \cite{bari2024allam}. Errors are especially common in factual recall and translation tasks, where linguistic complexity and syntactic differences further exacerbate hallucination rates. Recent efforts to address these challenges include integrating structured knowledge bases, utilizing Quranic and Classical Arabic corpora, applying retrieval-augmented generation (RAG), and adopting self-consistency decoding strategies. Combining RAG with Arabic MMLU and leveraging techniques like parameter-efficient fine-tuning and reinforcement learning from human feedback (RLHF) is anticipated to enhance both factual accuracy and dialectal robustness in Arabic LLMs \cite{bari2024allam,liang2024thames}.

\subsection{Model Compression Techniques}

As LLMs continue to expand, compression techniques like pruning and quantization have become crucial for enabling deployment on resource-constrained devices, though maintaining accuracy particularly for Arabic remains difficult. Structured pruning methods like FLOP and block-aware pruning reduce model size by removing less critical parameters while preserving dense structures and minimizing information loss \cite{wang2019structured}. Quantization strategies such as attention-aware PTQ, low-rank QAT, and hybrid quantization lower memory demands by reducing numerical precision, but aggressive quantization can introduce significant errors, especially in morphologically complex languages like Arabic \cite{yao2023posttraining} Research shows that compressed Arabic models, particularly after 4-bit quantization, experience higher hallucination rates, degraded factual recall, and weaker syntactic understanding compared to English models \cite{sibaee2024asos}. Looking ahead, it is crucial to develop compression methods specifically designed to preserve Arabic’s linguistic richness while balancing computational efficiency and factual reliability.

%%%%%%%%%%%%%%%%%%%%%%%%%%%%%%%%%%%%%%%%%%

\section{Methodology}

This study systematically evaluates the performance of six open-access LLMs — BLOOMZ, AceGPT, Jais, LLaMA-2, XGLM, and AraGPT — across three multilingual benchmarks: ArabicMMLU, EnglishMMLU, and Indic-Benchmark (Kannada-ARC-C-2.5K). The primary goals are to assess model adaptability to low-resource languages, particularly Arabic, and to analyze how compression techniques, including pruning and quantization, impact performance.

\subsection{Model Selection}

Model selection for this study was guided by the objective of capturing a broad spectrum of multilingual capabilities, architectural innovations, and training strategies relevant to low-resource languages. Jais and AraGPT were included due to their explicit focus on Arabic corpora, providing a baseline for evaluating performance on Arabic-language tasks. In contrast, BLOOMZ and XGLM were selected for their strong cross-lingual transfer abilities, making them ideal for assessing zero-shot generalization to Arabic and Indic benchmarks. Additionally, AceGPT and LLaMA-2, representing instruction-tuned models, were incorporated to examine whether enhanced prompt-following behavior improves performance on Arabic tasks without fine-tuning. Table \ref{tab:model_summary} summarizes all the chosen models with their configurations.

\begin{table*}[t]
    \centering
    \small
    \renewcommand{\arraystretch}{1.2}
    % Centred, automatically‐wrapping column type
    \newcolumntype{Y}{>{\centering\arraybackslash}X}
    \begin{tabularx}{\textwidth}{|c|c|Y|Y|}
        \hline
        \textbf{Model} & \textbf{Size} & \textbf{Specialization} & \textbf{Selection Rationale} \\ \hline
        BLOOMZ        & 560M / 7B      & Multilingual NLP, zero-shot learning     & Strong cross-lingual generalization          \\ \hline
        AceGPT        & 7B / 13B       & Conversational AI, instruction-tuned      & Evaluate Arabic instruction following        \\ \hline
        Jais          & 13B            & Arabic-centric NLP tasks                  & Arabic-optimised performance baseline        \\ \hline
        LLaMA-2       & 7B / 13B       & General NLP, chat models                  & General-purpose multilingual evaluation      \\ \hline
        XGLM          & 1.7B / 7.5B    & Multilingual NLP                          & Zero-shot transfer to low-resource languages \\ \hline
        AraGPT2       &  135M / 1.46B    & Arabic text generation                    & Benchmark Arabic-focused pre-training        \\ \hline
    \end{tabularx}
    \caption{Comprehensive overview of the language models evaluated in this study, outlining their parameter scale, domain specialisation, and the rationale for their inclusion in our experiments.}
    \label{tab:model_summary}
\end{table*}

In addition to language coverage, we prioritized architectural diversity in model selection. Although all selected models employ decoder-only Transformer architectures, they vary in optimization strategies, e.g., Grouped Query Attention (GQA) in LLaMA-2 and multitask fine-tuning in BLOOMZ and XGLM. A comparison of the smallest and largest members of these models permitted us to examine systematically the impact of architectural variation and model size on multilingual and Arabic-specific performance, particularly under compression techniques such as quantization and pruning. This permits one to see how architectural decisions affect generalization, specialization gains, and computational efficiency-accuracy-model size trade-offs under resource-constrained settings.

\subsection{Model Descriptions}

\textbf{BLOOMZ} is an extension of the BLOOM model, optimized for multilingual zero-shot learning and trained on 46 languages and 13 programming languages \cite{bloom2023}. While it excels at cross-lingual generalization, it demands high computational resources and struggles with long-form text consistency and biases.\\[0.5em]
\textbf{AceGPT} is an instruction-tuned, transformer-based model designed for multilingual text processing, dialogue systems, and code generation \cite{huang2024acegptlocalizinglargelanguage}. It emphasizes instruction-following, trained on diverse web, academic, and conversational datasets. Despite its strong dialogue performance, it is computationally expensive and prone to occasional hallucinations.\\[0.5em]
\textbf{Jais} is a 13B-parameter Arabic-first LLM developed for Arabic and English NLP tasks, specializing in text generation, translation, and summarization \cite{sengupta2023jais}. It uses a decoder-only architecture and is trained on 72 billion Arabic tokens plus multilingual data. Although highly effective for Arabic, it requires significant resources and shows bias toward Modern Standard Arabic over dialects.\\[0.5em]
\textbf{LLaMA 2} is a multilingual transformer-based model featuring GQA, SwiGLU activations, and pre-normalization \cite{touvron2023llama2openfoundation}. Trained on a carefully filtered web, academic, and code corpus, it supports strong performance in conversational AI and text generation tasks. However, it demands high-end GPUs and shares bias and long-form coherence limitations common to large models.\\[0.5em]
\textbf{XGLM} is a multilingual decoder-only LLM optimized for low-resource and cross-lingual text generation \cite{lin2022fewshotlearningmultilinguallanguage}. Trained on over 50 languages with multilingual tokenization, it achieves strong transferability across linguistic contexts but struggles with multilingual fairness and long-form coherence.\\[0.5em]
\textbf{AraGPT} is an Arabic-specialized LLM modeled after GPT-2 for Arabic text generation, translation, and sentiment analysis \cite{antoun2021aragpt2pretrainedtransformerarabic}. While it excels in Arabic generation, it faces challenges with dialect bias and long-form text generation.

\subsection{Datasets}

The \textbf{EnglishMMLU} benchmark \cite{hendrycks202engmmlu} evaluates a model's general knowledge across 57 subjects spanning STEM, social sciences, humanities, and specialized fields like law and medicine. It contains 15,908 manually collected questions sourced from exams and professional certifications, designed to assess real-world problem-solving rather than just linguistic ability. MMLU is a standard benchmark for measuring LLMs’ zero-shot and few-shot reasoning capabilities, revealing both the strengths of models in achieving human-level performance in many areas and their continued struggles with complex reasoning tasks.

\begin{figure}[htbp]
    \centering
    \includegraphics[width=0.6\linewidth]{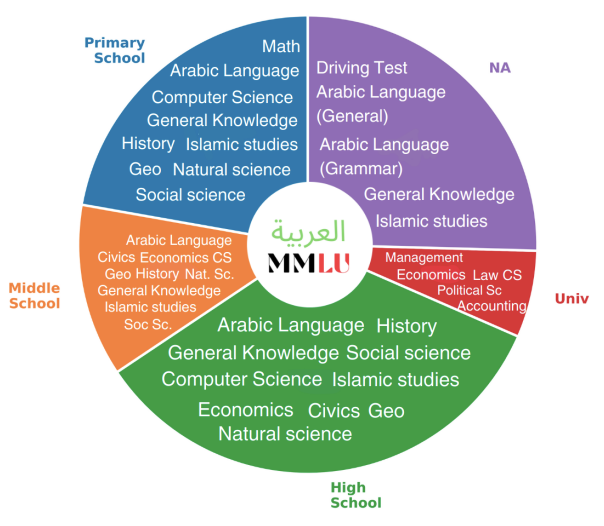} % 60% of column width
    \caption{ArabicMMLU Distribution of Educational Levels \& Corresponding Subjects \cite{koto2024arabicmmlu}}
    \label{fig:arabic}
\end{figure}

\textbf{ArabicMMLU} is a comprehensive benchmark designed to evaluate Arabic language models across 40 subjects and multiple educational levels, based on 14,575 multiple-choice questions drawn from eight Middle Eastern and North African countries \cite{koto2024arabicmmlu}. Covering disciplines from STEM to humanities, the dataset is structured across primary, middle school, high school, university, and professional levels, with careful quality control by native Arabic speakers as shown in Figure \ref{fig:arabic}. Despite its breadth, evaluations show that even leading models like GPT-4 achieve only 62.3\% accuracy, emphasizing the significant challenges and the critical role ArabicMMLU plays in advancing Arabic NLP research. 

The \textbf{IndicBenchmark} (Kannada-ARC-C-2.5K) \cite{kannada_arc_c_2_5k} dataset targets Kannada, a major Indic language, providing around 2,500 multiple-choice questions aimed at assessing comprehension, reasoning, and language understanding. Structured in Parquet format for accessibility, it addresses the underrepresentation of Indic languages in NLP research by offering a high-quality benchmark for low-resource language evaluation. Despite advances in multilingual modeling, Kannada remains a challenging language for state-of-the-art systems, highlighting the importance of benchmarks like Indic-Benchmark.

\subsection{Compression Techniques}

Weight pruning reduces the size and computational complexity of a neural network by removing weights with low importance. L1 unstructured pruning is applied to all linear layers, removing weights with the smallest absolute values. Pruning is performed at different levels (20\%, 40\%, 80\%) to assess the impact on performance. Given a weight matrix $W$ with elements $w_{ij}$, a threshold $\tau$ is chosen such that weights satisfying $|w_{ij}| < \tau$ are set to zero:

\begin{equation}
w_{ij} = 
\begin{cases}
0, & \text{if } |w_{ij}| < \tau \\
w_{ij}, & \text{otherwise}
\end{cases}
\end{equation}

Quantization reduces model precision to improve efficiency. By mapping floating-point weights to lower-bit representations (e.g., 4-bit and 8-bit integers), it reduces memory requirements and speeds up inference. The transformation follows: \begin{equation}
q = \operatorname{round}\left(\frac{w}{S}\right) + Z \end{equation} where $w$ is the original weight, $S$ is the scale factor, and $Z$ is the zero point ensuring zero representation in the quantized range. For 8-bit quantization, $q$ lies in $[0, 255]$ or $[-128, 127]$, while for 4-bit, $q$ ranges between $[0, 15]$ or $[-8, 7]$. The inverse transformation reconstructs weights as:

\begin{equation}
\hat{w} = S \cdot (q - Z)
\end{equation}

%%%%%%%%%%%%%%%%%%%%%%%%%%%%%%%%%%%%%%%%%%%

\section{Experimental Setup}

\noindent\textbf{Evaluation Pipeline.}
The pipeline consists of the following steps:
\begin{enumerate}[nosep,leftmargin=*,label=(\arabic*)]
    \item \textit{Model Initialization:} Load models and tokenizers with necessary configurations.
    \item \textit{Compression:} Apply pruning or quantization before inference.
    \item \textit{Prompt Design:} Tailor prompts per benchmark—ArabicMMLU uses metadata (subject, level, country), EnglishMMLU follows multiple-choice format, Indic-Benchmark uses Kannada prompts.
    \item \textit{Inference:} Extract logits at the final token position for each answer choice and select the answer with the highest probability.\\
\end{enumerate}

The performance metrics are accuracy (ratio of correct responses in all tasks) and confidence (mean softmax probability of the selected answer), and both provide an end-to-end measure of accuracy and certainty. Category-wise accuracy and confidence values are used to analyze fine-grained model performance in varying subject domains.\\

% \begin{equation}
% \text{Accuracy} = \left( \frac{\text{correct\_count}}{\text{total\_number\_of\_results}} \right) \times 100
% \end{equation}

% \begin{equation}
% \text{Category Accuracy} = \left( \frac{\text{number\_of\_correct\_predictions}}{\text{total\_examples\_in\_category}} \right) \times 100
% \end{equation}

% \begin{equation}
% \text{Confidence} = \max(\text{probabilities\_for\_options})
% \end{equation}

% \begin{equation}
% \text{Average Confidence} = \frac{\sum \text{Confidences}}{\text{total\_examples\_in\_category}}
% \end{equation}

Experiments were conducted on a Paperspace A100 GPU environment, with PyTorch and Hugging Face libraries supporting model handling, inference, and compression. All predictions, scores, and analysis results were systematically logged in structured CSV files to ensure full reproducibility.

%%%%%%%%%%%%%%%%%%%%%%%%%%%%%%%%%%%%%%%%%%%%%%%%%%%%%%%%%%%%%%%%%%%%%%%%
%%%%%%%%%%%%%%%%%%%%%%%%%%%%%%%%%%%%%%%%%%%%%%%%%%%%%%%%%%%%%%%%%%%%%%%%
%%%%%%%%%%%%%%%%%%%%%%%%%%%%%%%%%%%%%%%%%%%%%%%%%%%%%%%%%%%%%%%%%%%%%%%%

\section{Results}

This chapter offers an overall analysis of monolingual and multilingual LLMs for Arabic, English, and Indic Kannada benchmarks, investigating the effect of model capacity, pretraining in multiple languages, and compression strategies for accuracy, confidence, and cross-lingual generalization.

% \begin{figure}[h!]
%     \centering
%     \includegraphics[width=0.9\linewidth,height=5cm,keepaspectratio]{figures/arabicmmlu_model_conf.png} % Adjust width if needed
%     \caption{ArabicMMLU Model Categorical Confidence}
%     \label{fig:arabicmmluc_cat_accuracy}
% \end{figure}
\subsection{EnglishMMLU Evaluation}

Results for the EnglishMMLU benchmark (Table \ref{tab:engmmlu_model_acc}  establish the highest full-precision score for AceGPT-13B (47.7 \%), beating the multilingual BLOOMZ-7.1B (44.7 \%) and maintaining a considerable lead over all models with $\leq\!7$ B parameters (\,$\leq\!36.7$ \%). The trend confirms that large English instruction tuning supersedes parameter count alone (BLOOMZ vs. AceGPT-13B) or language specialisation alone (Jais-13B). Quantisation is essentially cost-free: converting any model to 8-bit—and for most of them, to 4-bit—alters accuracy by \mbox{$<\!2$\ \%}, with AceGPT-13B and BLOOMZ-7.1B even experiencing

Sparsification up to 20 \% is benign for all eleven systems (mean $\Delta\!=\!0.3$ pp). With 40 \% sparsity, accuracy starts to diverge: BLOOMZ-7.1B and AceGPT-13B fall only 3-4 \% but are still the leaders, while instruction-tuned AceGPT-7B drops 6.5\%. Extreme 80 \% sparsity brings the top-performers down: AceGPT-13B to 18.6 \%, BLOOMZ-7.1B to 24.0 \%-putting them at the level of small dense baselines like BLOOMZ-560M (23.0 \%). Therefore, making a strong model more compact brings obvious advantages only in the \emph{safe zone} ($\leq$ 40 \% sparsity); above that level the pre-training and depth advantages that characterize large models are effectively abolished. Further, we conducted an in-depth ablation study for each sub-category in the dataset, along with all models and configurations, for detailed insights as shown in Table \ref{tab:engmmlu_model_cat_acc}.

\begin{table}[h]
\centering
\scriptsize
\renewcommand{\arraystretch}{1.1}
\begin{tabular}{|l|c|c|c|c|c|c|}
\hline
\textbf{Model} & \textbf{FP} & \textbf{4Q} & \textbf{8Q} & \textbf{20\%P} & \textbf{40\%P} & \textbf{80\%P} \\
\hline
BLOOMZ-560M & 25.97 & 25.93 & 26.23 & 25.80 & 25.11 & 22.95 \\
BLOOMZ-7.1B & 44.67 & 43.81 & 44.72 & 44.59 & 41.58 & 24.04 \\
AceGPT-7B & 36.72 & 23.53 & 23.86 & 31.46 & 30.27 & 24.87 \\
AceGPT-13B & 47.74 & 48.12 & 47.78 & 46.09 & 43.34 & 18.61 \\
Jais-13B & 25.77 & 24.12 & 23.54 & 25.74 & 24.78 & 23.91 \\
LLaMA 2-7B & 27.20 & 25.85 & 25.82 & 27.75 & 26.81 & 21.53 \\
LLaMA 2-13B & 30.61 & 29.93 & 26.32 & 30.84 & 29.84 & 24.46 \\
XGLM-1.7B & 22.92 & 22.92 & 22.96 & 23.55 & 15.38 & 26.89 \\
XGLM-7.5B & 23.27 & 22.92 & 23.10 & 22.97 & 24.64 & 24.28 \\
AraGPT2-135M & 23.84 & 20.18 & 23.63 & 23.88 & 23.07 & 21.09 \\
AraGPT2-1.47B & 9.60 & 24.12 & 9.69 & 5.66 & 1.38 & 14.82 \\
\hline
\end{tabular}
\caption{EnglishMMLU Model Accuracy}
\label{tab:engmmlu_model_acc}
\end{table}
% \vspace{0.5em} % optional: small space between table and figure
% \begin{figure}[h!]
%     \centering
%     \includegraphics[width=0.9\linewidth,height=5cm,keepaspectratio]{figures/engmmlu_cat_acc.png}
%     \caption{EnglishMMLU Model Categorical Accuracy}
%     \label{fig:engmmlu_cat_acc}
% \end{figure}

\subsection{ArabicMMLU Evaluation} 
Table~\ref{tab:arabicmmlu_model_acc} shows the results of ArabicMMLU on all categories together for different settings.
Across all 11 evaluated systems, larger generalist models consistently outperform their smaller or expert-specialized counterparts. Notably, BLOOMZ-7.1B attains the highest full-precision (FP) accuracy at 41.7\%, followed closely by the instruction-tuned AceGPT-13B (40.3\%) and the Arabic-specialized Jais-13B ({36.0\%}). In contrast, all models with $\leq$7B parameters fall below 34\%. This disparity highlights that scale and broad multilingual pretraining—especially when combined with multitask supervision—offer greater leverage for Arabic reasoning than pure language-specific pretraining (e.g., Jais vs. BLOOMZ) or scale alone (e.g., AceGPT vs. BLOOMZ). Moreover, quantization incurs minimal loss, reducing to 8-bit—and in most cases 4-bit—precision leads to a drop of less than 2\% points in accuracy, suggesting that low-precision inference should be the default for resource-constrained deployment.

Pruning reveals a clear accuracy-efficiency trade-off consistent across model families. Up to 20\% sparsity yields negligible accuracy degradation ($<1$ \%). At 40\% sparsity, model-dependent effects emerge: BLOOMZ-7.1B and Jais-13B remain strong at 36.9\% and 38.8\%, respectively—still outperforming all dense models under 7B—whereas AceGPT-13B and other instruction-tuned variants drop sharply by 8–11\%, likely due to reduced redundancy in task-specialized layers. Under extreme 80\% pruning, accuracy drops substantially across the board, yet a pruned BLOOMZ-7.1B (effectively $\sim$1.4B parameters) still matches or exceeds dense 1–2B baselines. This validates that \textit{compressing a strong, large model is generally preferable to using a natively small one}. In summary: (i) favor the strongest multilingual model available, (ii) apply aggressive quantization, and (iii) prune moderately, ideally capping sparsity. 

Overall performance across ArabicMMLU is consistently weaker than across EnglishMMLU, which is a dramatic cross-lingual gap caused by an imbalance in the amount of available data. Even Arabic fine-tuned versions underperform because there is much less volume and quality of Arabic data than for English. The consequence is ongoing cultural and linguistic bias towards English, which distorts the reasoning abilities of the models in Arabic domains. To gain a clearer insight into the nature of this behaviour, we performed a detailed ablation across all subject domains in the ArabicMMLU set, as listed in Table~\ref{tab:arabicmmlu_cat_acc}.

\begin{table}[h]
\centering
\scriptsize
\renewcommand{\arraystretch}{1.2}
\begin{tabular}{|l|c|c|c|c|c|c|}
\hline
\textbf{Model} & \textbf{FP} & \textbf{4Q} & \textbf{8Q} & \textbf{20\%P} & \textbf{40\%P} & \textbf{80\%P} \\
\hline
BLOOMZ-560M & 27.02 & 27.46 & 26.70 & 26.43 & 31.12 & 29.36 \\
BLOOMZ-7.1B & 41.71 & 39.86 & 41.86 & 40.79 & 36.88 & 30.56 \\
AceGPT-7B & 34.22 & 32.93 & 32.00 & 29.52 & 26.76 & 21.20 \\
AceGPT-13B & 40.27 & 39.16 & 40.59 & 36.98 & 29.69 & 21.16 \\
Jais-13B & 36.01 & 35.59 & 35.03 & 35.68 & 38.79 & 32.06 \\
LLaMA 2-7B & 31.27 & 31.20 & 31.26 & 28.79 & 30.47 & 25.89 \\
LLaMA 2-13B & 31.38 & 30.94 & 31.22 & 32.49 & 28.43 & 22.25 \\
XGLM-1.7B & 30.90 & 28.17 & 30.95 & 29.29 & 19.74 & 21.14 \\
XGLM-7.5B & 31.24 & 31.16 & 31.20 & 31.20 & 28.75 & 26.47 \\
AraGPT2-135M & 31.04 & 30.47 & 30.95 & 31.06 & 31.14 & 25.89 \\
AraGPT2-1.47B & 31.15 & 31.18 & 31.15 & 31.16 & 31.17 & 27.77 \\
\hline
\end{tabular}
\caption{ArabicMMLU Model Accuracy}
\label{tab:arabicmmlu_model_acc}
\end{table}

\subsection{Indic Benchmark (Kannada) Evaluation}
\begin{figure}[t!]
    \centering
    \includegraphics[width=\linewidth]{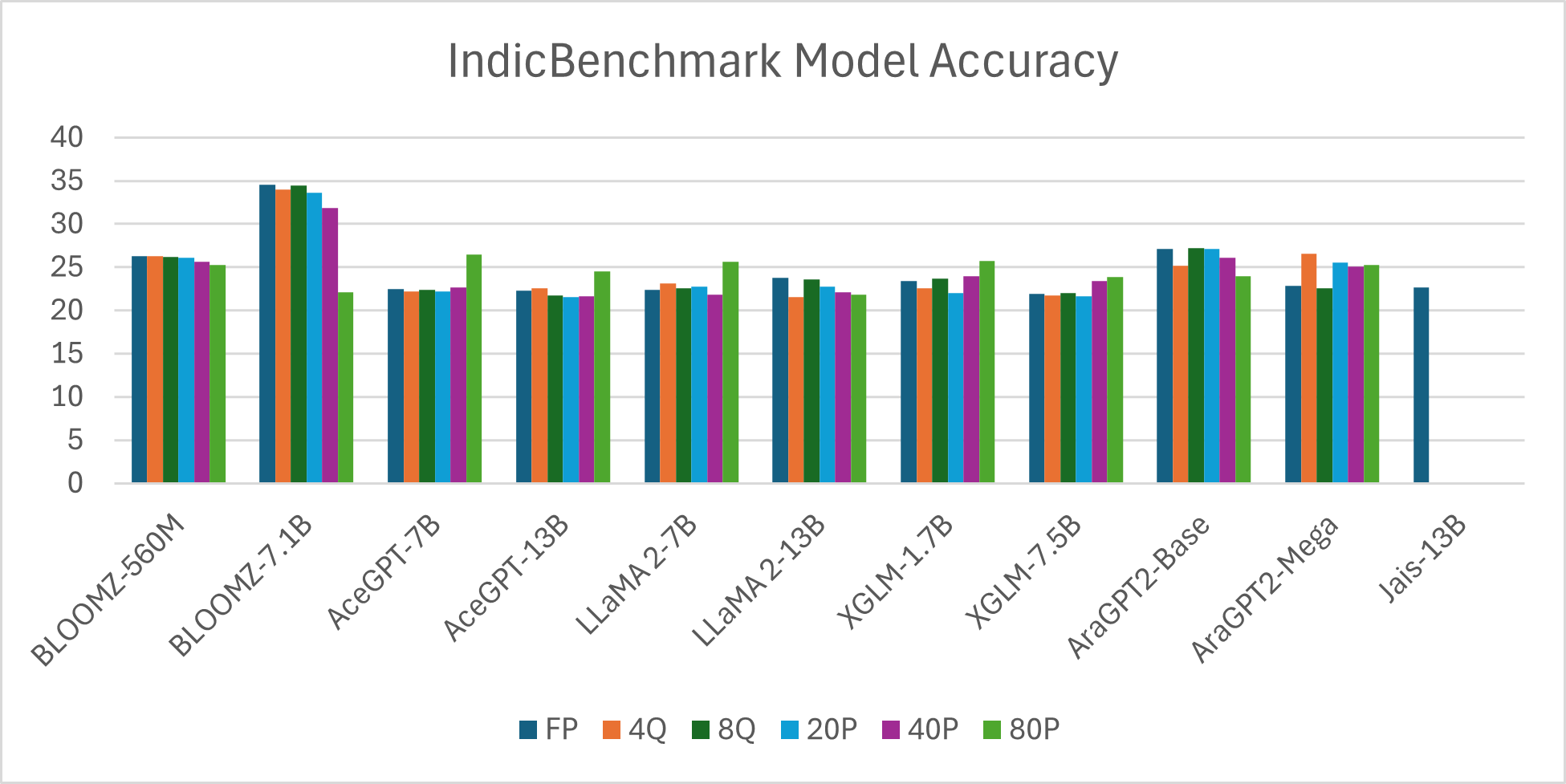} % Adjust width if needed
    \caption{IndicBenchmark Models' Accuracy Comparison in Different Configurations}
    \label{fig:indic_model_accuracy}
\end{figure}

Table~\ref{tab:indic_acc} and Figure~\ref{fig:indic_model_accuracy} present results on the Indic Kannada benchmark. BLOOMZ-7.1B led with 34.56\% full-precision accuracy, followed by BLOOMZ-560M (26.32\%). AceGPT and XGLM variants scored slightly lower, reflecting low-resource adaptation challenges. Quantization to 8-bit precision maintained accuracy within 0.3\%, confirming its safety as a compression method. However, 80\% pruning caused significant declines for all models, particularly larger ones.

The accuracy spread between full-precision and 8-bit quantized models remains under 0.3\%, underscoring quantization’s minimal impact even in scarce-data settings. Monolingual models like AraGPT2-1.47B underperform multilingual counterparts by 12\%, indicating the necessity of dedicated adaptation strategies such as synthetic data augmentation. Pruning-induced cross-lingual drift is evident: at 40\% sparsity, BLOOMZ-7.1B’s accuracy drops by 7.9\% on Kannada, compared to 4.8\% on Arabic and 3.1\% on English, suggesting that cross-lingual representations for low-resource languages are less redundantly encoded and more sensitive to weight removal.

Indic-Kannada shows the worst performance among the three languages that we scored, lagging much behind both Arabic and English. This is because it is a very low-resource language with very little presence in the pretraining corpora of most of the best-performing models. In contrast, English is supported by large quantities of such training data, and Arabic is the one for which other models are specifically fine-tuned. Kannada does not have any such targeted large-scale training or fine-tuning. Therefore, as a consequence, the models are not generalizable. Also, since subject-wise subcategories are not included in the Indic-Benchmark, we couldn't carry out a detailed ablation study like those for the Arabic and English languages.

\begin{table}[h]
\centering
\scriptsize
\renewcommand{\arraystretch}{1.1}
\begin{tabular}{|l|c|c|c|c|c|c|}
\hline
\textbf{Model} & \textbf{FP} & \textbf{4Q} & \textbf{8Q} & \textbf{20\%P} & \textbf{40\%P} & \textbf{80\%P} \\
\hline
BLOOMZ-560M & 26.32 & 26.24 & 26.20 & 26.08 & 25.64 & 25.25 \\
BLOOMZ-7.1B & 34.56 & 34.01 & 34.48 & 33.65 & 31.83 & 22.12 \\
AceGPT-7B & 22.43 & 22.16 & 22.39 & 22.20 & 22.63 & 26.48 \\
AceGPT-13B & 22.31 & 22.59 & 21.76 & 21.52 & 21.64 & 24.53 \\
Jais-13B & 22.63 & 22.01 & 21.68 & 21.97 & 20.56 & 22.34 \\
LLaMA 2-7B & 22.39 & 23.11 & 22.55 & 22.79 & 21.84 & 25.60 \\
LLaMA 2-13B & 23.78 & 21.56 & 23.62 & 22.71 & 22.08 & 21.80 \\
XGLM-1.7B & 23.38 & 22.59 & 23.66 & 22.00 & 23.94 & 25.76 \\
XGLM-7.5B & 21.88 & 21.76 & 22.00 & 21.60 & 23.38 & 23.82 \\
AraGPT2-135M & 27.15 & 25.17 & 27.23 & 27.15 & 26.12 & 23.98 \\
AraGPT2-1.47B & 22.83 & 26.60 & 22.59 & 25.49 & 25.09 & 25.25 \\
\hline
\end{tabular}
\caption{IndicBenchmark Model Accuracy}
\label{tab:indic_acc}
\end{table}

\subsection{Cross-Benchmark Comparative Analysis}
Comparative analysis across ArabicMMLU, EnglishMMLU, and Indic Kannada benchmarks shows consistent trends influenced by model size, diversity in training, and availability of training material. Multilingual models, such as BLOOMZ, consistently outperform with the generalization encouraged by pretraining over broad language sets, which helps in effective generalization. AceGPT performs well over EnglishMMLU owing to English exclusive training, but degrades over Indic Kannada owing to training data differences. Monolingual models such as AraGPT2 perform well over ArabicMMLU but do not generalize well across benchmarks, attesting to the disadvantage of language specialization.

Compression techniques reveal other trends: 4-bit and 8-bit quantization are very lossless, decreasing accuracy by around 2\% across all benchmarks. Pruning has trade-offs—sparsity above 40\% degrades performance heavily, particularly for large models. Such degradation tends to undermine accuracy improvements over denser, smaller counterparts. Overall, performance tracks the resource gradient of the data: \texttt{EnglishMMLU} $>$ \texttt{ArabicMMLU} $>$ \texttt{Indic Kannada}, with strong predictive confidence vs. accuracy correlations. In conclusion, the insights are:

\begin{itemize}

\item \textbf{Increase in English-Arabic Performance Gap} The accuracy difference between large and small models is much wider in EnglishMMLU than in ArabicMMLU. This implies that the abundance of English data enables large models to learn general linguistic patterns, while the lack of Arabic data constrains size-based improvements.
\item \textbf{Sensitivity of Larger Models to Pruning} Pruning to 20–40\% sparsity incurs negligible accuracy losses (<3\%), whereas pruning above 40\% can make pruned large models perform worse than unpruned smaller counterparts. The phenomenon suggests that aggressive weight elimination destroys the most essential network structure for fine-grained reasoning.
\item \textbf{Multilingual Pretraining vs. Monolingual Specialization} Multilingual models like BLOOMZ-7.1B outperform monolingual Arabic models such as AraGPT2-1.47B on ArabicMMLU, demonstrating that shared cross-lingual embeddings built during multilingual pretraining improve robustness and generalization, whereas monolingual specialization may lead to overfitting narrow domains.
\end{itemize}

%%%%%%%%%%%%%%%%%%%%%%%%%%%%%%%%%%%%%%%%%%%%%%%%%%%%%%%%%%%%%%%%%%%%%%%%
%%%%%%%%%%%%%%%%%%%%%%%%%%%%%%%%%%%%%%%%%%%%%%%%%%%%%%%%%%%%%%%%%%%%%%%%
%%%%%%%%%%%%%%%%%%%%%%%%%%%%%%%%%%%%%%%%%%%%%%%%%%%%%%%%%%%%%%%%%%%%%%%%

%%%%%%%%%%%%%%%%%%%%%%%%%%%%%%%%%%%%%%%%%%%%%%%%%%%%%%%%%%%%%%%%%%%%%%%%
%%%%%%%%%%%%%%%%%%%%%%%%%%%%%%%%%%%%%%%%%%%%%%%%%%%%%%%%%%%%%%%%%%%%%%%%
%%%%%%%%%%%%%%%%%%%%%%%%%%%%%%%%%%%%%%%%%%%%%%%%%%%%%%%%%%%%%%%%%%%%%%%%

\section{Limitations}

Though this paper presents valuable insights into both monolingual and multilingual LLM performance in Arabic, English, and Kannada, it is constrained by some limitations. Firstly, experiments only included zero-shot scenarios and did not test the impact of a few-shot setting. Secondly, pruning and quantization strategies were utilized, yet sophisticated strategies like movement pruning and knowledge distillation were not considered. Thirdly, the evaluation was only conducted with MMLU benchmarks, not including human-human annotated tasks or actual applications, and the analysis of hallucinations was not systematic. Lastly, the investigation covered only Modern Standard Arabic and academic Kannada, ignoring the dialectal differences and the domain of code-switching phenomena, both of which are promising avenues for future investigation.

\section{Future Work}

Future work would aim to extend benchmark datasets to more low-resource languages like Arabic and Kannada, with a priority on developing high-quality datasets that reflect dialect variability and genuine usage context. Systematic evaluation towards mitigating hallucination and bias is also important with the use of advanced fine-tuning, targeted prompt engineering, and bias-correction approaches. Optimization of LLM efficiency using advanced compression techniques like movement pruning, knowledge distillation, and enhanced quantization is also important in enabling their deployment in resource-scarce settings. Collaboration across disciplines involving linguistics, cognitive science, AI ethics, and human-in-the-loop approaches would significantly increase the reliability, contextual awareness, and general trust of multilingual LLMs.

\section{Recommendations}

There are a number of recommended key strategies to improve LLM performance on low-resource languages. First, using modular multilingual architectures can make a big difference by dynamically routing the input to language-specific models. Second, the use of synthetic data augmentation—back-translating high-resource datasets into low-resource languages—can significantly increase the size of training corpora and the ability of the model to generalize. Third, the evaluation frameworks can transition from fixed scholarly yardsticks to dynamic, interactive testing that more closely reflects real-world usage and user behaviors. The adoption of these strategies will make the development of multilingual AI systems more scalable, robust, and inclusive.

%%%%%%%%%%%%%%%%%%%%%%%%%%%%%%%%%%%%%%%%%%%%%%%%%%%%%%%%%%%%%%%%%%%%%%%%
%%%%%%%%%%%%%%%%%%%%%%%%%%%%%%%%%%%%%%%%%%%%%%%%%%%%%%%%%%%%%%%%%%%%%%%%
%%%%%%%%%%%%%%%%%%%%%%%%%%%%%%%%%%%%%%%%%%%%%%%%%%%%%%%%%%%%%%%%%%%%%%%%

\section{Conclusion}
This work provides a structured comparison of multilingual and monolingual LLMs on Arabic, English, and Indic (Kannada) benchmarks, characterizing the accuracy–efficiency trade-offs brought by compression. Cross-lingual robustness is stronger in multilingual models, as demonstrated by BLOOMZ-7.1B, than in language-specialized counterparts. Four-bit and eight-bit quantization reduced memory and latency at minimal loss of accuracy, beneficial to hardware-restricted deployments, while moderate sparsity was sufficient before noticeable loss of stability and performance, particularly in larger models. Towards advancing the community, we propose dynamic, interaction-oriented benchmarking, dialect-aware enriching of corpora, principled hallucination reduction, and modular multilingual adapters. These directions collectively offer more resilient, inclusive, and scalable NLP solutions.

\include{engmmlu_arammlu_table}

\bibliographystyle{splncs04}

\bibliography{mybibfile}
\end{document}

%% file: engmmlu_arammlu_table.tex
\begin{table*}[h!]
\centering
\scriptsize
\renewcommand{\arraystretch}{1.1}
\centering
\resizebox{\textwidth}{!}{
\begin{tabularx}{\textwidth}{|p{4cm}|XXXX|}
\hline
\textbf{Model Name} & \textbf{STEM} & \textbf{Human\newline -ities} & \textbf{Social \newline Sciences} & \textbf{Other} \\
\hline
BLOOMZ-560M & 26.58 & 28.32 & 24.80 & 24.83 \\
\hline
BLOOMZ-560M-4Q & 26.69 & 27.97 & 24.19 & 25.11 \\
\hline
BLOOMZ-560M-8Q & 26.55 & 28.35 & 24.59 & 25.55 \\
\hline
BLOOMZ-560M-20P & 27.01 & 28.48 & 24.02 & 24.43 \\
\hline
BLOOMZ-560M-40P & 25.88 & 26.22 & 23.80 & 24.64 \\
\hline
BLOOMZ-560M-80P & 21.31 & 23.77 & 21.27 & 23.96 \\
\hline
BLOOMZ-7.1B & 40.35 & 42.04 & 51.44 & 45.56 \\
\hline
BLOOMZ-7.1B-4Q & 39.22 & 41.19 & 50.87 & 44.71 \\
\hline
BLOOMZ-7.1B-8Q & 40.39 & 42.01 & 51.88 & 45.51 \\
\hline
BLOOMZ-7.1B-20P & 40.46 & 42.20 & 51.22 & 45.32 \\
\hline
BLOOMZ-7.1B-40P & 37.17 & 39.47 & 46.77 & 42.84 \\
\hline
BLOOMZ-7.1B-80P & 22.87 & 25.90 & 23.41 & 23.82 \\
\hline
AceGPT-7B & 32.71 & 37.58 & 36.72 & 38.21 \\
\hline
AceGPT-7B-4Q & 22.19 & 24.24 & 22.27 & 24.29 \\
\hline
AceGPT-7B-8Q & 22.37 & 24.77 & 22.75 & 24.54 \\
\hline
AceGPT-7B-20P & 29.81 & 32.56 & 29.52 & 32.45 \\
\hline
AceGPT-7B-40P & 26.41 & 31.59 & 29.61 & 31.71 \\
\hline
AceGPT-7B-80P & 26.09 & 25.59 & 23.89 & 24.26 \\
\hline
AceGPT-13B & 39.50 & 47.63 & 53.49 & 49.56 \\
\hline
AceGPT-13B-4Q & 40.14 & 49.01 & 52.79 & 49.74 \\
\hline
AceGPT-13B-8Q & 40.14 & 47.91 & 52.93 & 49.15 \\
\hline
AceGPT-13B-20P & 37.10 & 47.22 & 50.09 & 48.29 \\
\hline
AceGPT-13B-40P & 36.46 & 44.46 & 47.82 & 44.32 \\
\hline
AceGPT-13B-80P & 23.86 & 8.82 & 21.70 & 20.22 \\
\hline
Jais-13B & 24.67 & 25.87 & 25.20 & 26.49 \\
\hline
Jais-13B-20P & 28.04 & 28.45 & 30.26 & 29.50 \\
\hline
Jais-13B-40P & 25.95 & 27.10 & 29.08 & 27.69 \\
\hline
Jais-13B-80P & 22.97 & 24.02 & 22.71 & 24.80 \\
\hline
LLaMA 2-7B & 26.76 & 30.08 & 25.15 & 26.63 \\
\hline
LLaMA 2-7B-4Q & 26.83 & 27.82 & 23.71 & 25.13 \\
\hline
LLaMA 2-7B-8Q & 26.48 & 28.04 & 23.45 & 25.20 \\
\hline
Llama-2-7b-hf-20P & 27.08 & 29.32 & 25.11 & 26.56 \\
\hline
LLaMA 2-7B-40P & 27.26 & 28.23 & 26.16 & 26.07 \\
\hline
LLaMA 2-7B-80P & 21.88 & 25.43 & 20.52 & 19.59 \\
\hline
LLaMA 2-13B & 27.12 & 32.31 & 30.35 & 31.49 \\
\hline
LLaMA 2-13B-4Q & 27.47 & 31.49 & 29.56 & 30.42 \\
\hline
LLaMA 2-13B-8Q & 24.71 & 28.38 & 24.98 & 26.51 \\
\hline
LLaMA 2-13B-20P & 26.48 & 29.20 & 28.95 & 30.22 \\
\hline
LLaMA 2-13B-40P & 28.32 & 30.80 & 27.95 & 27.53 \\
\hline
LLaMA 2-13B-80P & 24.53 & 24.08 & 25.72 & 24.14 \\
\hline
XGLM-1.7B & 21.24 & 23.80 & 21.40 & 23.86 \\
\hline
XGLM-1.7B-4Q & 21.73 & 23.61 & 21.18 & 23.82 \\
\hline
XGLM-1.7B-8Q & 21.38 & 23.89 & 21.27 & 23.96 \\
\hline
XGLM-1.7B-20P & 22.09 & 24.33 & 22.14 & 24.38 \\
\hline
XGLM-1.7B-40P & 13.98 & 17.36 & 18.17 & 15.33 \\
\hline
XGLM-1.7B-80P & 27.75 & 23.92 & 31.09 & 26.33 \\
\hline
XGLM-7.5B & 22.23 & 24.18 & 21.57 & 23.95 \\
\hline
XGLM-7.5B-4Q & 21.31 & 23.67 & 21.31 & 23.95 \\
\hline
XGLM-7.5B-8Q & 21.59 & 24.27 & 21.53 & 23.82 \\
\hline
XGLM-7.5B-20P & 21.52 & 23.83 & 21.62 & 23.89 \\
\hline
XGLM-7.5B-40P & 22.58 & 23.99 & 25.68 & 25.18 \\
\hline
XGLM-7.5B-80P & 23.68 & 24.30 & 22.71 & 25.18 \\
\hline
AraGPT2-135M & 23.86 & 25.21 & 22.84 & 23.46 \\
\hline
AraGPT2-135M-4Q & 19.82 & 21.82 & 18.52 & 20.11 \\
\hline
AraGPT2-135M-8Q & 23.40 & 25.40 & 22.45 & 23.23 \\
\hline
AraGPT2-135M-20P & 24.14 & 25.21 & 23.14 & 23.30 \\
\hline
AraGPT2-135M-40P & 21.27 & 24.08 & 21.83 & 23.89 \\
\hline
AraGPT2-135M-80P & 20.67 & 17.65 & 20.61 & 23.41 \\
\hline
AraGPT2-1.47B & 6.87 & 12.72 & 8.25 & 9.75 \\
\hline
AraGPT2-1.47B-4Q & 25.63 & 24.93 & 23.01 & 23.37 \\
\hline
AraGPT2-1.47B-8Q & 7.04 & 12.40 & 8.47 & 9.96 \\
\hline
AraGPT2-1.47B-20P & 4.25 & 10.74 & 5.63 & 6.46 \\
\hline
AraGPT2-1.47B-40P & 2.51 & 4.11 & 3.89 & 3.50 \\
\hline
AraGPT2-1.47B-80P & 15.96 & 18.96 & 11.22 & 13.39 \\
\hline
\end{tabularx}
}
\caption{EnglishMMLU Model Categorical Accuracy}
\label{tab:engmmlu_model_cat_acc}
\end{table*}

\begin{table*}[h!]
\centering
\scriptsize
\renewcommand{\arraystretch}{1.1}

\centering
\resizebox{\textwidth}{!}{
\begin{tabularx}{\textwidth}{|p{4cm}|XXXXX|}
\hline
\textbf{Model Name} & \textbf{STEM} & \textbf{Human\newline -ities} & \textbf{Social \newline Science} & \textbf{Language} & \textbf{Other} \\
\hline
BLOOMZ-560M & 28.39 & 27.74 & 26.53 & 24.02 & 26.89 \\
\hline
BLOOMZ-560M-4Q & 29.22 & 28.24 & 26.10 & 24.56 & 27.93 \\
\hline
BLOOMZ-560M-8Q & 27.92 & 27.17 & 26.16 & 24.44 & 26.69 \\
\hline
BLOOMZ-560M-20P & 27.98 & 27.06 & 25.73 & 24.02 & 26.09 \\
\hline
BLOOMZ-560M-40P & 30.12 & 32.48 & 30.23 & 25.89 & 35.17 \\
\hline
BLOOMZ-560M-80P & 29.57 & 29.77 & 30.48 & 28.06 & 27.77 \\
\hline
BLOOMZ-7.1B & 39.88 & 41.67 & 44.58 & 34.98 & 44.54 \\
\hline
BLOOMZ-7.1B-4Q & 38.35 & 39.34 & 42.63 & 33.29 & 43.02 \\
\hline
BLOOMZ-7.1B-8Q & 39.97 & 42.00 & 44.63 & 35.34 & 44.50 \\
\hline
BLOOMZ-7.1B-20P & 38.73 & 40.57 & 43.39 & 34.02 & 44.58 \\
\hline
BLOOMZ-7.1B-40P & 36.49 & 34.69 & 39.72 & 32.63 & 39.38 \\
\hline
BLOOMZ-7.1B-80P & 28.98 & 33.82 & 29.69 & 24.44 & 33.13 \\
\hline
AceGPT-7B & 33.48 & 35.70 & 31.55 & 31.31 & 38.70 \\
\hline
AceGPT-7B-4Q & 32.48 & 33.84 & 30.62 & 30.52 & 37.05 \\
\hline
AceGPT-7B-8Q & 32.80 & 36.03 & 31.50 & 30.64 & 37.82 \\
\hline
AceGPT-7B-20P & 31.46 & 26.95 & 29.07 & 26.37 & 33.53 \\
\hline
AceGPT-7B-40P & 28.94 & 24.19 & 25.23 & 24.02 & 31.69 \\
\hline
AceGPT-7B-80P & 21.43 & 18.85 & 21.84 & 24.38 & 21.33 \\
\hline
AceGPT-13B & 38.79 & 43.09 & 39.66 & 32.63 & 44.02 \\
\hline
AceGPT-13B-4Q & 36.27 & 42.71 & 38.76 & 32.15 & 42.90 \\
\hline
AceGPT-13B-8Q & 39.66 & 43.15 & 40.37 & 33.11 & 43.34 \\
\hline
AceGPT-13B-20P & 35.09 & 37.95 & 37.46 & 31.19 & 41.18 \\
\hline
AceGPT-13B-40P & 30.78 & 27.55 & 30.79 & 26.97 & 31.65 \\
\hline
AceGPT-13B-80P & 21.71 & 20.52 & 20.85 & 24.86 & 19.37 \\
\hline
Jais-13B & 33.88 & 37.18 & 33.90 & 30.52 & 43.66 \\
\hline
Jais-13B-20P & 33.29 & 36.99 & 33.79 & 30.34 & 43.06 \\
\hline
Jais-13B-40P & 36.37 & 41.34 & 36.58 & 32.03 & 45.78 \\
\hline
Jais-13B-80P & 33.20 & 32.72 & 30.06 & 27.39 & 35.57 \\
\hline
LLaMA 2-7B & 30.68 & 31.82 & 30.68 & 27.45 & 34.61 \\
\hline
LLaMA 2-7B-4Q & 30.12 & 26.48 & 30.03 & 22.76 & 28.61 \\
\hline
LLaMA 2-7B-8Q & 30.59 & 31.87 & 30.54 & 27.51 & 34.73 \\
\hline
Llama-2-7b-hf-20P & 27.58 & 31.52 & 28.25 & 26.37 & 28.73 \\
\hline
LLaMA 2-7B-40P & 29.94 & 30.73 & 29.69 & 27.39 & 33.93 \\
\hline
LLaMA 2-7B-80P & 26.21 & 26.84 & 24.27 & 19.02 & 30.93 \\
\hline
LLaMA 2-13B & 30.06 & 31.74 & 30.28 & 28.36 & 36.09 \\
\hline
LLaMA 2-13B-4Q & 29.41 & 31.63 & 29.97 & 28.00 & 35.25 \\
\hline
LLaMA 2-13B-8Q & 29.88 & 31.63 & 30.14 & 28.18 & 35.93 \\
\hline
LLaMA 2-13B-20P & 32.42 & 32.37 & 30.62 & 28.96 & 37.78 \\
\hline
LLaMA 2-13B-40P & 28.70 & 26.59 & 30.20 & 26.19 & 29.73 \\
\hline
LLaMA 2-13B-80P & 23.14 & 21.23 & 22.09 & 23.18 & 22.21 \\
\hline
XGLM-1.7B & 29.88 & 31.11 & 30.37 & 27.93 & 34.61 \\
\hline
XGLM-1.7B-4Q & 29.32 & 25.69 & 29.52 & 23.90 & 31.25 \\
\hline
XGLM-1.7B-8Q & 29.78 & 31.19 & 30.37 & 28.06 & 34.85 \\
\hline
XGLM-1.7B-20P & 28.60 & 28.54 & 29.63 & 24.38 & 34.05 \\
\hline
XGLM-1.7B-40P & 19.29 & 18.63 & 20.06 & 19.02 & 21.97 \\
\hline
XGLM-1.7B-80P & 22.30 & 19.78 & 20.65 & 23.18 & 20.97 \\
\hline
XGLM-7.5B & 30.34 & 31.57 & 30.59 & 28.18 & 34.85 \\
\hline
XGLM-7.5B-4Q & 29.81 & 31.55 & 30.54 & 28.24 & 35.17 \\
\hline
XGLM-7.5B-8Q & 30.34 & 31.55 & 30.54 & 28.30 & 34.65 \\
\hline
XGLM-7.5B-20P & 30.16 & 31.46 & 30.45 & 28.30 & 35.13 \\
\hline
XGLM-7.5B-40P & 26.40 & 29.44 & 27.32 & 28.72 & 32.85 \\
\hline
XGLM-7.5B-80P & 28.51 & 23.64 & 28.62 & 22.58 & 27.53 \\
\hline
AraGPT2-135M & 30.16 & 31.49 & 30.11 & 27.93 & 34.89 \\
\hline
AraGPT2-135M-4Q & 29.81 & 30.45 & 29.75 & 27.63 & 34.25 \\
\hline
AraGPT2-135M-8Q & 30.00 & 31.41 & 30.14 & 27.93 & 34.65 \\
\hline
AraGPT2-135M-20P & 29.84 & 31.55 & 30.31 & 27.87 & 35.09 \\
\hline
AraGPT2-135M-40P & 29.88 & 31.46 & 30.54 & 28.24 & 35.05 \\
\hline
AraGPT2-135M-80P & 23.07 & 25.06 & 27.40 & 26.73 & 28.05 \\
\hline
AraGPT2-1.47B & 30.16 & 31.16 & 30.54 & 28.00 & 35.37 \\
\hline
AraGPT2-1.47B-4Q & 29.88 & 31.55 & 30.54 & 28.24 & 35.17 \\
\hline
AraGPT2-1.47B-8Q & 30.25 & 31.35 & 30.31 & 28.06 & 35.25 \\
\hline
AraGPT2-1.47B-20P & 29.91 & 31.55 & 30.40 & 28.06 & 35.33 \\
\hline
AraGPT2-1.47B-40P & 29.91 & 31.55 & 30.48 & 28.18 & 35.21 \\
\hline
AraGPT2-1.47B-80P & 27.58 & 30.78 & 28.36 & 25.83 & 24.05 \\
\hline
\end{tabularx}
}
\caption{ArabicMMLU Model Categorical Accuracy}
\label{tab:arabicmmlu_cat_acc}
\end{table*}

%% file: samplepaper.bbl
\begin{thebibliography}{10}
\providecommand{\url}[1]{\texttt{#1}}
\providecommand{\urlprefix}{URL }
\providecommand{\doi}[1]{https://doi.org/#1}

\bibitem{almazrouei2023alghafa}
Almazrouei, E., Cojocaru, R., Baldo, M., Malartic, Q., Alobeidli, H., Mazzotta, D., Penedo, G., Campesan, G., Farooq, M., Alhammadi, M., et~al.: Alghafa evaluation benchmark for arabic language models. In: Proceedings of ArabicNLP 2023. pp. 244--275 (2023)

\bibitem{antoun2021aragpt2pretrainedtransformerarabic}
Antoun, W., Baly, F., Hajj, H.: Aragpt2: Pre-trained transformer for arabic language generation. arXiv preprint arXiv:2012.15520  (2020)

\bibitem{Sindhu2024evolutionllm}
B, S., P, P.R., B, S.M., S, K.: The evolution of large language model: Models, applications and challenges pp.~1--8 (2024). \doi{10.1109/ICCTAC61556.2024.10581180}

\bibitem{bari2024allam}
Bari, M.S., Alnumay, Y., Alzahrani, N.A., Alotaibi, N.M., Alyahya, H.A., AlRashed, S., Mirza, F.A., Alsubaie, S.Z., Alahmed, H.A., Alabduljabbar, G., et~al.: Allam: Large language models for arabic and english. arXiv preprint arXiv:2407.15390  (2024)

\bibitem{boughorbel2023analyzing}
Boughorbel, S., Hawasly, M.: Analyzing multilingual competency of llms in multi-turn instruction following: A case study of arabic. arXiv preprint arXiv:2310.14819  (2023)

\bibitem{cheung2024reality}
Cheung, M.: A reality check of the benefits of llm in business. arXiv preprint arXiv:2406.10249  (2024)

\bibitem{chu2025llm}
Chu, Z., Wang, S., Xie, J., Zhu, T., Yan, Y., Ye, J., Zhong, A., Hu, X., Liang, J., Yu, P.S., et~al.: Llm agents for education: Advances and applications. arXiv preprint arXiv:2503.11733  (2025)

\bibitem{dey2024better}
Dey, K., Tarannum, P., Hasan, M.A., Razzak, I., Naseem, U.: Better to ask in english: Evaluation of large language models on english, low-resource and cross-lingual settings. arXiv preprint arXiv:2410.13153  (2024)

\bibitem{Hagos2024}
Hagos, D.H., Battle, R., Rawat, D.B.: Recent advances in generative ai and large language models: Current status, challenges, and perspectives. IEEE Transactions on Artificial Intelligence  \textbf{5}(12),  5873--5893 (2024). \doi{10.1109/TAI.2024.3444742}

\bibitem{hasan2024large}
Hasan, M.A., Tarannum, P., Dey, K., Razzak, I., Naseem, U.: Do large language models speak all languages equally? a comparative study in low-resource settings. arXiv preprint arXiv:2408.02237  (2024)

\bibitem{hendrycks202engmmlu}
Hendrycks, D., Burns, C., Basart, S., Zou, A., Mazeika, M., Song, D., Steinhardt, J.: Measuring massive multitask language understanding. arXiv preprint arXiv:2009.03300  (2020)

\bibitem{hijazi2024arablegaleval}
Hijazi, F., AlHarbi, S., AlHussein, A., Shairah, H.A., AlZahrani, R., AlShamlan, H., Knio, O., Turkiyyah, G.: Arablegaleval: A multitask benchmark for assessing arabic legal knowledge in large language models. arXiv preprint arXiv:2408.07983  (2024)

\bibitem{huang2024acegptlocalizinglargelanguage}
Huang, H., Yu, F., Zhu, J., Sun, X., Cheng, H., Song, D., Chen, Z., Alharthi, A., An, B., He, J., et~al.: Acegpt, localizing large language models in arabic. arXiv preprint arXiv:2309.12053  (2023)

\bibitem{kannada_arc_c_2_5k}
{Indic-Benchmark}: Indic-benchmark/kannada-arc-c-2.5k. \url{https://huggingface.co/datasets/Indic-Benchmark/kannada-arc-c-2.5k} (2023)

\bibitem{koto2024arabicmmlu}
Koto, F., Li, H., Shatnawi, S., Doughman, J., Sadallah, A.B., Alraeesi, A., Almubarak, K., Alyafeai, Z., Sengupta, N., Shehata, S., et~al.: Arabicmmlu: Assessing massive multitask language understanding in arabic. arXiv preprint arXiv:2402.12840  (2024)

\bibitem{liang2024thames}
Liang, M., Arun, A., Wu, Z., Munoz, C., Lutch, J., Kazim, E., Koshiyama, A., Treleaven, P.: Thames: An end-to-end tool for hallucination mitigation and evaluation in large language models. arXiv preprint arXiv:2409.11353  (2024)

\bibitem{lin2022fewshotlearningmultilinguallanguage}
Lin, X.V., Mihaylov, T., Artetxe, M., Wang, T., Chen, S., Simig, D., Ott, M., Goyal, N., Bhosale, S., Du, J., et~al.: Few-shot learning with multilingual language models. arXiv preprint arXiv:2112.10668  (2021)

\bibitem{nafea2024araoffence}
Nafea, Y., Shehata, S., Talat, Z., Aboeitta, A., Sharshar, A., Nakov, P.: Araoffence: Detecting offensive speech across dialects in arabic media. In: Proceedings of Interspeech 2024. pp. 4303--4307 (2024). \doi{10.21437/Interspeech.2024-2077}

\bibitem{narayanan2024suvach}
Narayanan, V., KP, P.R., Nouphal, S.: Suvach--generated hindi qa benchmark. arXiv preprint arXiv:2404.19254  (2024)

\bibitem{nazi2024large}
Nazi, Z.A., Peng, W.: Large language models in healthcare and medical domain: A review. In: Informatics. vol.~11, p.~57. MDPI (2024)

\bibitem{plaza2024spanish}
Plaza, I., Melero, N., del Pozo, C., Conde, J., Reviriego, P., Mayor-Rocher, M., Grandury, M.: Spanish and llm benchmarks: is mmlu lost in translation? arXiv preprint arXiv:2406.17789  (2024)

\bibitem{refai2023data}
Refai, D., Abu-Soud, S., Abdel-Rahman, M.J.: Data augmentation using transformers and similarity measures for improving arabic text classification. IEEE Access  \textbf{11},  132516--132531 (2023)

\bibitem{sengupta2023jais}
Sengupta, N., Sahu, S.K., Jia, B., Katipomu, S., Li, H., Koto, F., Marshall, W., Gosal, G., Liu, C., Chen, Z., et~al.: Jais and jais-chat: Arabic-centric foundation and instruction-tuned open generative large language models. arXiv preprint arXiv:2308.16149  (2023)

\bibitem{sibaee2024asos}
Sibaee, S.T., Alharbi, A.I., Ahmed, S., Nacar, O., Ghouti, L., Koubaa, A.: Asos at arabic llms hallucinations 2024: Can llms detect their hallucinations. In: Proceedings of the 6th Workshop on Open-Source Arabic Corpora and Processing Tools (OSACT) with Shared Tasks on Arabic LLMs Hallucination and Dialect to MSA Machine Translation@ LREC-COLING 2024. pp. 130--134 (2024)

\bibitem{singh2024indicgenbench}
Singh, H., Gupta, N., Bharadwaj, S., Tewari, D., Talukdar, P.: Indicgenbench: a multilingual benchmark to evaluate generation capabilities of llms on indic languages. arXiv preprint arXiv:2404.16816  (2024)

\bibitem{taghanaki2024mmlu}
Taghanaki, S.A., Khani, A., Khasahmadi, A.: Mmlu-pro+: Evaluating higher-order reasoning and shortcut learning in llms. arXiv preprint arXiv:2409.02257  (2024)

\bibitem{touvron2023llama2openfoundation}
Touvron, H., Martin, L., Stone, K., Albert, P., Almahairi, A., Babaei, Y., Bashlykov, N., Batra, S., Bhargava, P., Bhosale, S., et~al.: Llama 2: Open foundation and fine-tuned chat models. arXiv preprint arXiv:2307.09288  (2023)

\bibitem{wang2019structured}
Wang, Z., Wohlwend, J., Lei, T.: Structured pruning of large language models. arXiv preprint arXiv:1910.04732  (2019)

\bibitem{bloom2023}
Workshop, B., Scao, T.L., Fan, A., Akiki, C., Pavlick, E., Ili{\'c}, S., Hesslow, D., Castagn{\'e}, R., Luccioni, A.S., Yvon, F., et~al.: Bloom: A 176b-parameter open-access multilingual language model. arXiv preprint arXiv:2211.05100  (2022)

\bibitem{yao2023posttraining}
Yao, Z., Li, C., Wu, X., Youn, S., He, Y.: A comprehensive study on post-training quantization for large language models. arXiv preprint  \textbf{arXiv:2303.08302} (2023)

\end{thebibliography}
